\documentclass[11pt,twocolumn]{article}

\usepackage{booktabs} 

\usepackage[utf8]{inputenc}

\usepackage{authblk}

\usepackage{amsmath}
\usepackage{amssymb}

\usepackage{mathtools}

\usepackage{graphicx}
\usepackage{epstopdf}

\usepackage{url}

\usepackage{algorithm}
\usepackage{xcolor}
\usepackage{listings}
\usepackage[export]{adjustbox}

\newcommand{\ram}{\ensuremath{\texttt{RAM}}{}}
\newcommand{\cbf}{\ensuremath{\texttt{CBF}}{}}

\newcommand{\ed}{\ensuremath{\texttt{ED}}{}}
\newcommand{\dtw}{\ensuremath{\texttt{DTW}}{}}
\newcommand{\erp}{\ensuremath{\texttt{ERP}}{}}

\newcommand{\frechet}{\ensuremath{\texttt{DK}}{}}

\renewcommand{\leq}{\leqslant}

\renewcommand{\phi}{\ensuremath{\varphi}}
\renewcommand{\epsilon}{\ensuremath{\varepsilon}}

\newcount\colveccount
\newcommand*\colvec[1]{
        \global\colveccount#1
        \begin{pmatrix}
        \colvecnext
}
\def\colvecnext#1{
        #1
        \global\advance\colveccount-1
        \ifnum\colveccount>0
                \\
                \expandafter\colvecnext
        \else
                \end{pmatrix}
        \fi
}

\begin{document}

\title{High Dimensional Time Series Generators}

\author[1]{J\"org P. Bachmann}
\author[2]{Johann-Christoph Freytag}
\affil[1]{ \texttt{joerg.bachmann@informatik.hu-berlin.de}}
\affil[2]{ \texttt{freytag@informatik.hu-berlin.de}}
\affil[1,2]{ Humboldt-Universität zu Berlin, Germany}
\date{\today}

\maketitle

\begin{abstract}
    Multidimensional time series are sequences of real valued vectors.
    They occur in different areas, for example handwritten characters, GPS tracking, and gestures of modern virtual reality motion controllers.
    Within these areas, a common task is to search for similar time series.
    Dynamic Time Warping (\dtw) is a common distance function to compare two time series.
    The Edit Distance with Real Penalty (\erp) and the Dog Keeper Distance (\frechet) are two more distance functions on time series.
    Their behaviour has been analyzed on 1-dimensional time series.
    However, it is not easy to evaluate their behaviour in relation to growing dimensionality.
    For this reason we propose two new data synthesizers generating multidimensional time series.
    The first synthesizer extends the well known cylinder-bell-funnel (CBF) dataset to multidimensional time series.
    Here, each time series has an arbitrary type (cylinder, bell, or funnel) in each dimension, thus for $d$-dimensional time series there are $3^{d}$ different classes.
    The second synthesizer (\ram) creates time series with ideas adapted from Brownian motions which is a common model of movement in physics.
    Finally, we evaluate the applicability of a 1-nearest neighbor classifier using \dtw{} on datasets generated by our synthesizers.
\end{abstract}

\section{Introduction}
\label{sec:introduction}

\begin{figure*}
    \includegraphics[width=.3\linewidth,valign=T]{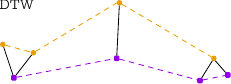}
    \hfill
    \includegraphics[width=.3\linewidth,valign=T]{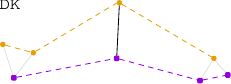}
    \hfill
    \includegraphics[width=.3\linewidth,valign=T]{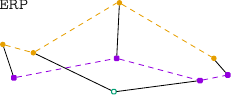}
    \caption{
        Examples of aligning two time series using \dtw{} (left), \frechet{} (center), and \erp{} (right).
        Distances between states are marked with solid lines while the circled and squared time series are connected using dashed lines.
        The green blank circle is the gap element for \erp{}.
    }
    \label{fig:sketchDTW}
    \label{fig:sketchDK}
    \label{fig:sketchERP}
\end{figure*}

Multimedia retrieval is a common application which requires finding similar time series to a given query.
This includes gesture recognition with modern virtual reality motion controllers, GPS tracking, speech recognition, and classification of handwritten letters.
In these areas, time series of the same classes (e.\,g., same written character or same gestures) follow the same path in space, but have some temporal displacements.
Tracking the GPS coordinates of two cars driving the same route from A to B is another example.
We want these tracks (i.\,e. the \emph{trajectories of the time series}) to be recognized as similar, although driving style, traffic lights, and traffic jams might result in temporal differences.
Distance functions such as dynamic time warping (\dtw) \cite{DTWSakoe}, edit distance with real penalties (\erp) \cite{ERP}, and the dog-keeper distance (\frechet) \cite{computingfrechet} respect this semantic requirement.
These \emph{time warping distance functions} map pairs of time series representing similar trajectories to small distances and dissimilar time series to large distances.
They basically follow the same idea by finding good alignments of the elements between the two elements (cf. Figure~\ref{fig:sketchDTW} for examples).

Of course, we are interested in fast algorithms for these time series distance functions.
Unfortunately, time warping distance functions usually have quadratic runtime \cite{FrechetNoSubQuadratic,DTWNoSubQuadratic}.
There are some evaluations of different time warping distance functions on different data sets \cite{TSComparison}.
However, their performance has not been evaluated in relation to growing dimensionality.

\paragraph*{Computation time of distance function:}
In order to evaluate the computation time of a distance function in relation to growing dimensionality, we need datasets with similar properties (e.\,g. size of dataset, length of time series, data distribution) but different dimensionality.
Existing datasets of different dimensionality exist, but have different properties.
Thus, datasets with similar properties but different dimensionalities are necessary to evaluate relations between the computation time and the dimensionality.
To achieve similar properties on datasets with different dimensionality, we propose the usage of synthesized data.

\paragraph*{Tightness of lower bounds:}
Nearest neighbor queries are also accelerated by pruning distance computations using cheap lower bounds \cite{ExactIndexingDTW,Trillion}.
If a lower bound claims a large distance then there is no need to compute the exact but expensive distance value.
The lower bound proposed by Keogh was extended to multidimensional time series in an unpublished paper \cite{LBKeoghMultivariat} but there is no evaluation available regarding growing dimensionality.
Again, having datasets with similar properties for different dimensionality is necessary for these evaluations.
Those datasets could also be used to evaluate pruning strategies of index structures specialized to \dtw{} \cite{IndexingDTWWithED} or metric index structures \cite{covertree,mvptree,mtree,mindex} applicable for metric time series distances (e.\,g. \frechet{} \cite{WDK} and \erp \cite{ERP}).

\paragraph*{Accuracy of classifiers:}
Nearest neighbor queries appear in supervised machine learning, for example 1-nearest neighbor classifiers.
In order to compare properties of two classifiers using two distinct distance functions, labeled datasets are necessary.
Since the comparison is meaningless when both classifier are perfectly accurate, we need to control the difficulty of the classification tasks.
Thus, when designing dataset generators for the purpose of comparing the strength of two classifiers, we need to implement a parameter to control the difficulty.

\paragraph*{Time distortion of trajectories:}
All of our proposed evaluation tasks consider time warping distance functions which yields the demand for time distorting dataset generators.
Otherwise, evaluations would prefer implementations of distance functions which have an advantage on comparing perfectly aligned time series.
The results of those evaluations could not be transfered to datasets with time distorted time series.

\paragraph*{Contribution:}
To make such evaluation scenarios feasable, we propose two dataset generators for multidimensional, labeled, and time distorted time series.
Both generators provide a tuning parameter to control the difficulty of generated classification tasks.

The rest of the paper is structured as follows.
Section~\ref{sec:cbf} presents the first dataset generator which is an extension of the well known cylinder-bell-funnel dataset \cite{CBF}.
Section~\ref{sec:ram} presents the \ram{} dataset generator which adapts ideas from Brownian motions.
We evaluate the datasets using \dtw{} in Section~\ref{sec:evaluation} in order to confirm that \dtw{} yields an applicable 1-nearest neighbor classifier and that we can control its classification score.

\section{Cylinder-Bell-Funnel}
\label{sec:cbf}

N. Saito proposed the well known 1-dimensional cylinder-bell-funnel dataset in his PhD thesis \cite{CBF}.
It is an artificial dataset consisting of three different classes of time series: cylinder, bell, and funnel.

For the time series synthesizer, let $\ell>0$ be a fixed length of the time series and $\mathcal N$ be a standard normal distributed random variable.
Furthermore, fixiate $a$ and $b$ uniformly distributed over $\left[ \ell\cdot\frac 1 8, \ell\cdot\frac 2 8 \right]$ and $\left[ \ell\cdot\frac 6 8, \ell\cdot\frac 7 8  \right]$, respectively and $\nu=6+\mathcal N$ per generated time series.
Each time series has a prefix $\mathcal P$ of length $a$ and suffix $\mathcal S$ of length $\ell-b$ containing standard normal distributed random numbers $\left( \mathcal N, \cdots, \mathcal N \right)$.

\begin{figure}
    \centering
    \includegraphics[width=.32\linewidth]{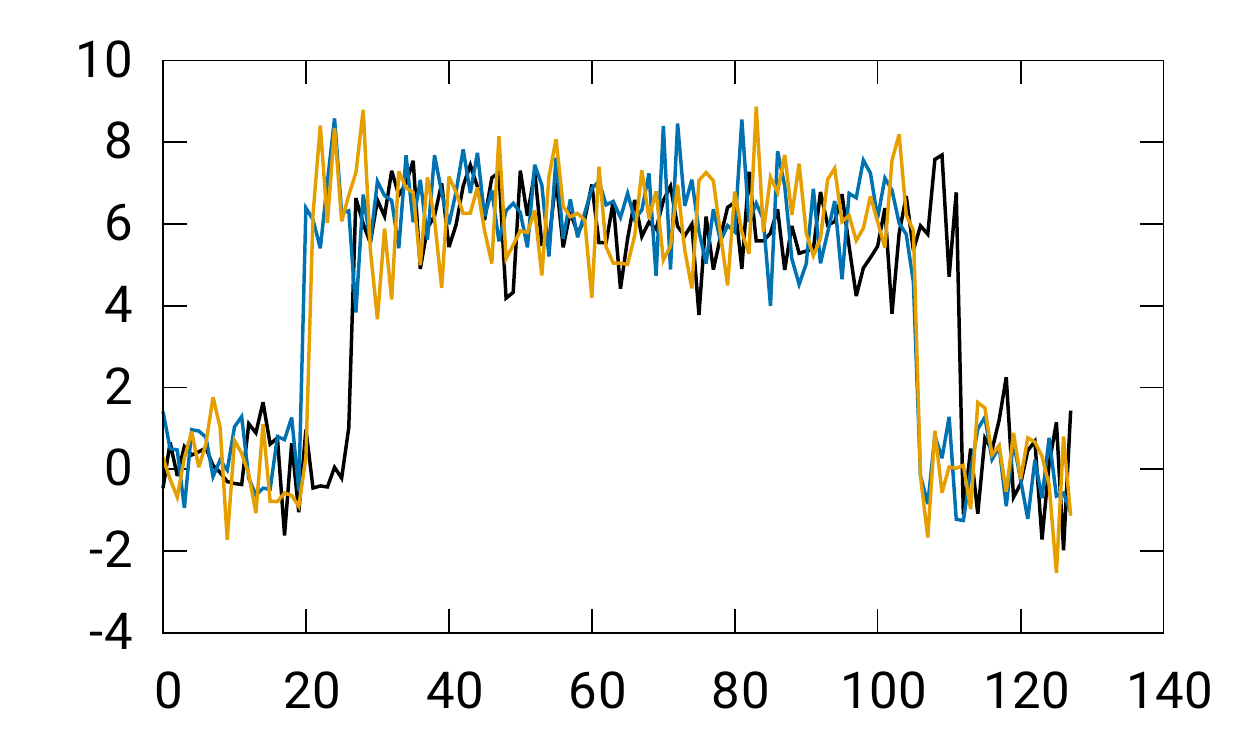}
    \includegraphics[width=.32\linewidth]{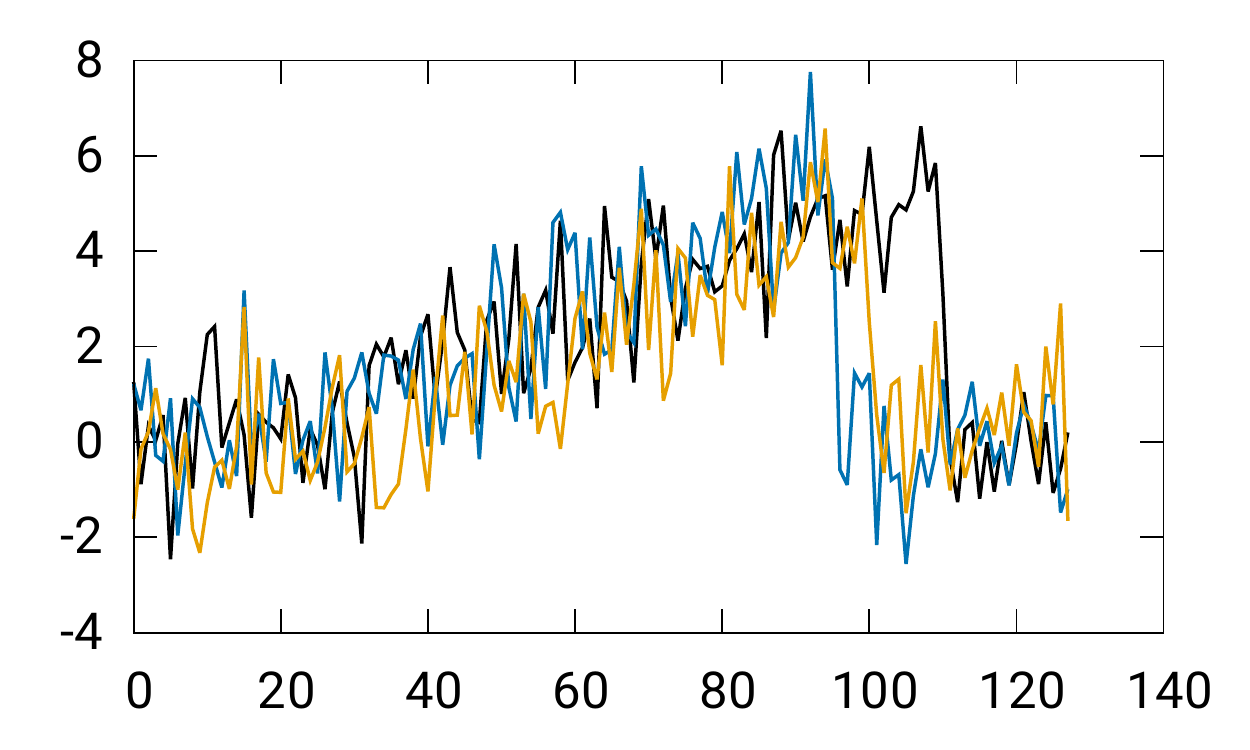}
    \includegraphics[width=.32\linewidth]{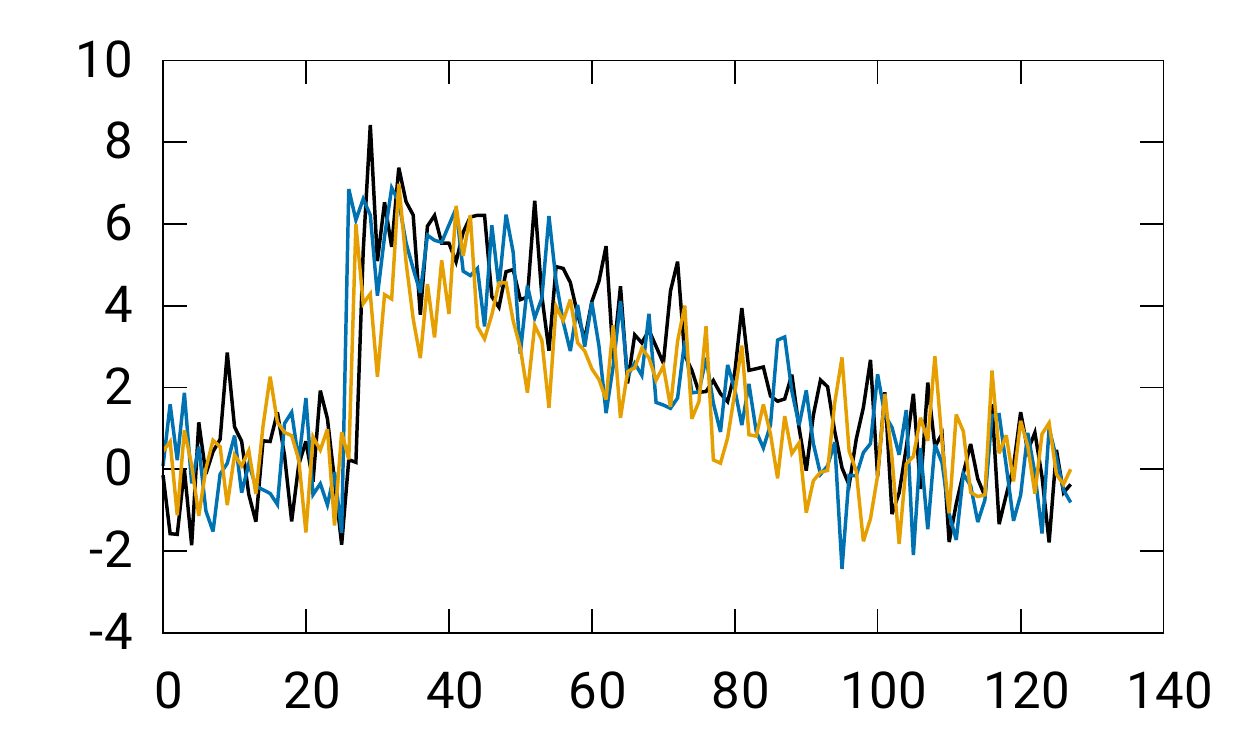}
    \caption{Examples for cylinder (left), bell (center), and funnel (right) time series resp..}
    \label{fig:cbf}
\end{figure}

The middle parts of random \emph{cylinder} ($\mathcal C$), \emph{bell} ($\mathcal B$) and \emph{funnel} ($\mathcal F$) time series are a plateau, a rising linear function, and a falling linear function, respectively.
Their length is $b-a$, thus
\begin{align}
    \label{eq:cbf}
    \mathcal C &\coloneqq \mathcal P\times\left( \cdots, \nu+\mathcal N, \cdots \right)\times\mathcal S \\
    \mathcal B &\coloneqq \mathcal P\times\left( \cdots, \nu\cdot\frac{i-a}{b-a}+\mathcal N, \cdots \right)\times\mathcal S \\
    \mathcal F &\coloneqq \mathcal P\times\left( \cdots, \nu\cdot\frac{b-i}{b-a}+\mathcal N, \cdots \right)\times\mathcal S
\end{align}
where $\nu=6+\mathcal N$ is chosen once per time series (cf. Figure~\ref{fig:cbf} for examples).

\begin{figure}
    \centering
    \includegraphics[width=.32\linewidth]{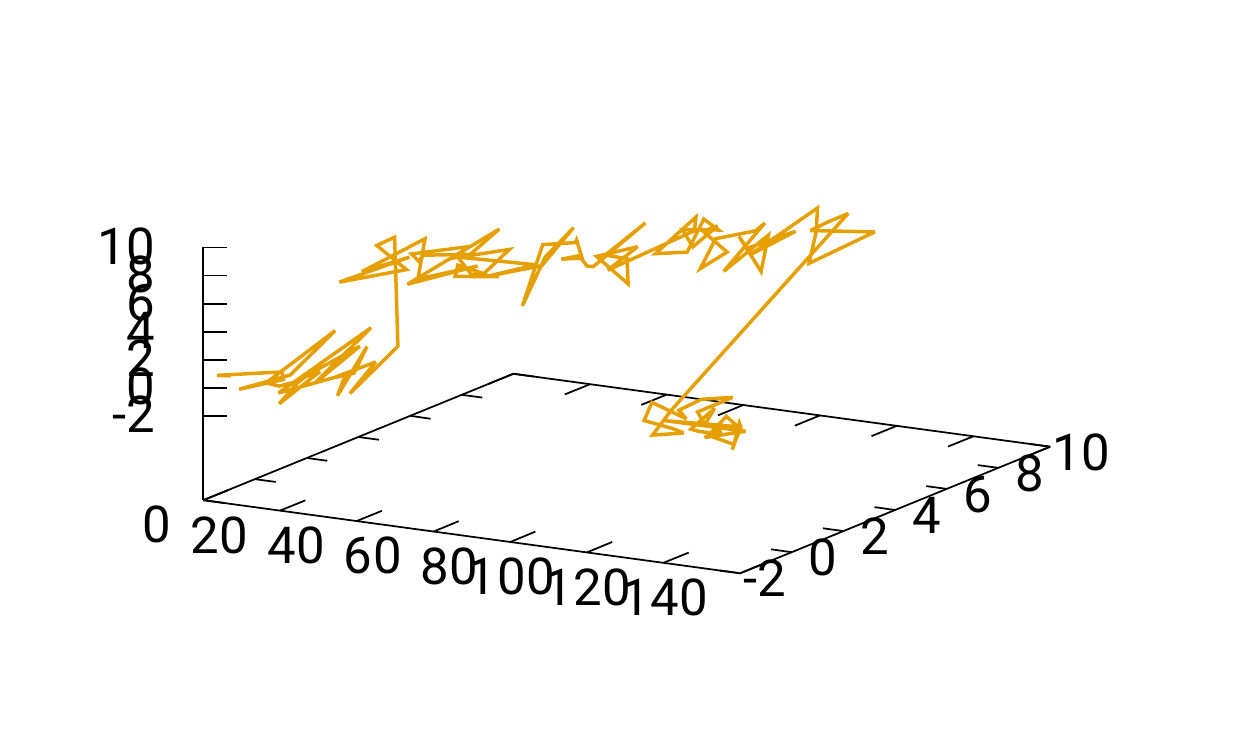}
    \includegraphics[width=.32\linewidth]{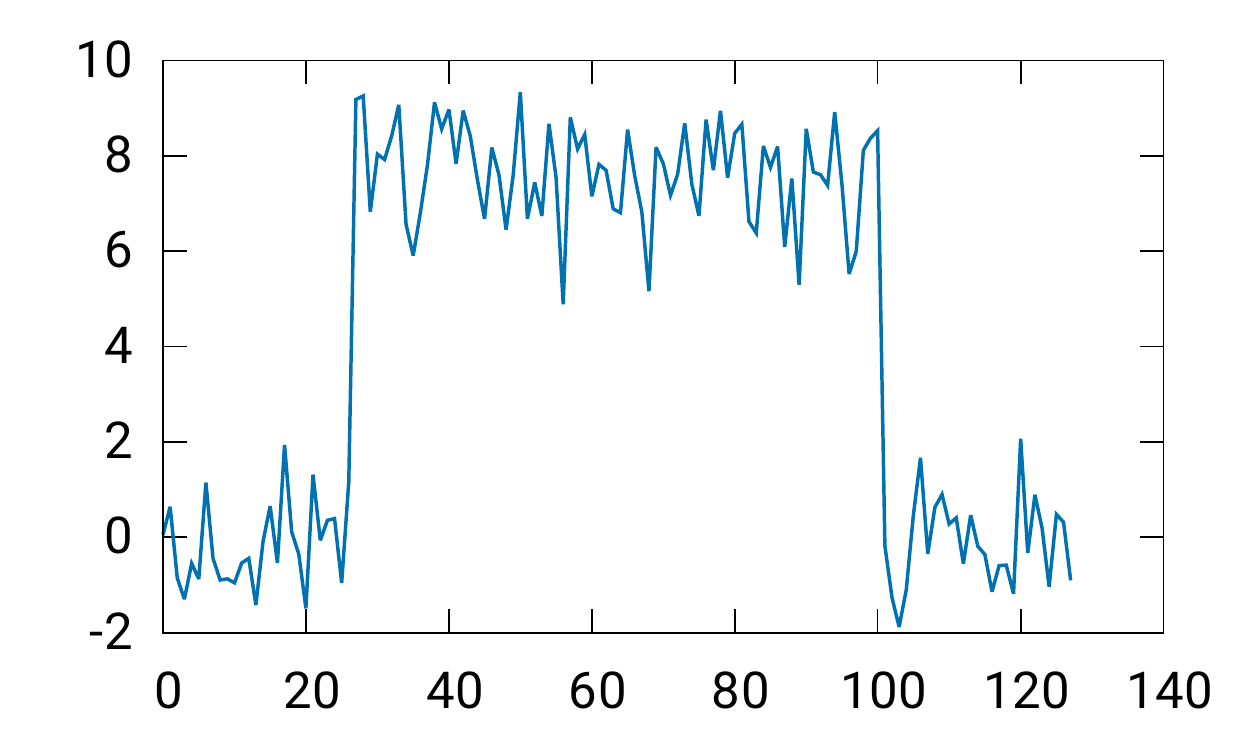}
    \includegraphics[width=.32\linewidth]{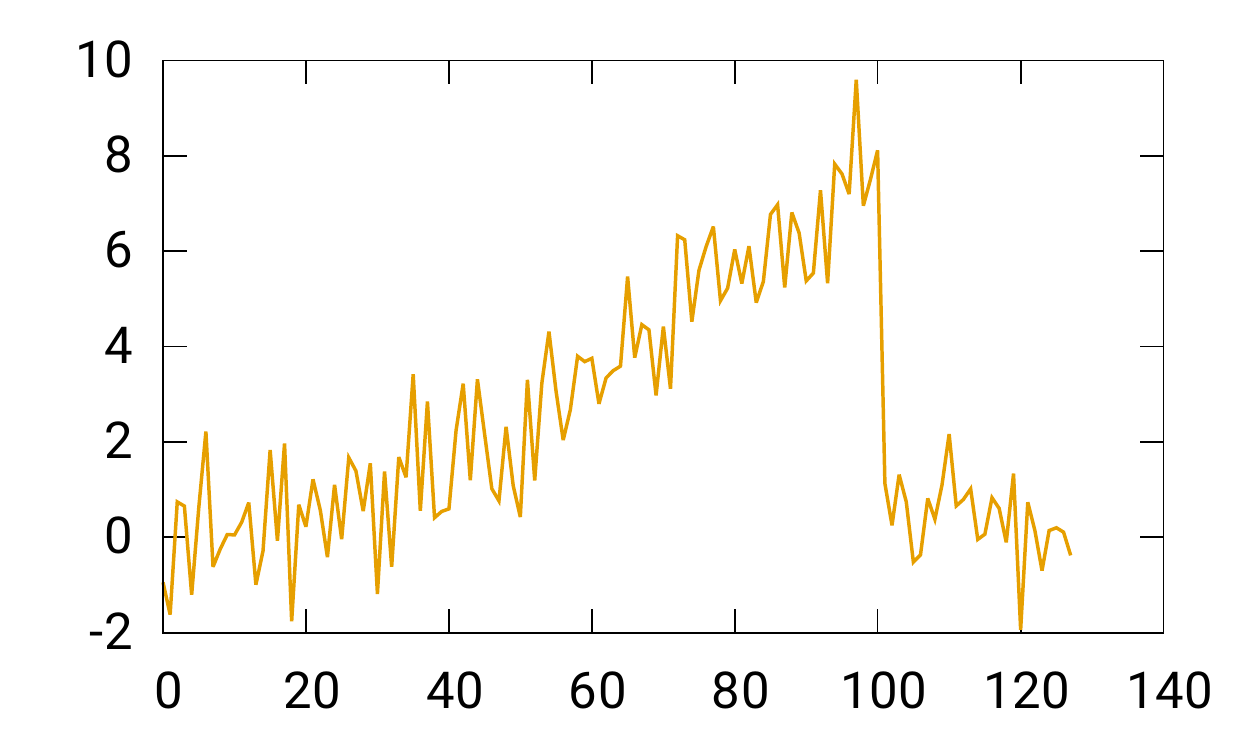}
    \caption{An example for a 2-dimensional CBF time series (left).
    Its first dimension is a cylinder (center) and its second dimension is a bell (right).}
    \label{fig:cbf_md}
\end{figure}

We canonically extend \cbf{} to generate multidimensional time series by generating one of these types for each dimension with the same starting and ending positions $a$ and $b$ (cf. Figure~\ref{fig:cbf_md} for a 2-dimensional example).
Thus, the input parameters are the length of the time series and a vector $\mathbf t$ with values $c$, $b$, and $f$ claiming which type to synthesize per dimension.
Given the dimensionality $n$, the multidimensional \cbf{} generator produces a maximum number of $3^n$ different classes, which is the set of all combinations of $c$, $b$, and $f$.

\section{Random Accelerated Motion Generator}
\label{sec:ram}

The \ram{} generator produces classes by first generating \emph{base time series} using impulse driven motions with random acceleration (cf. Section~\ref{sec:basetimeseries} for details).
\emph{Representatives} of the classes are generated by distorting the base time series in space and time.

\subsection{Base Time Series Generator}
\label{sec:basetimeseries}

The Brownian motion is a common model of random motion in physics, for example to model movement of molecules in gases.
Basically, each next position of a molecule is obtained as a randomized position around the current position.

In order to achieve more \textit{curve like} time series, we keep an impulse vector and add that to the current position to obtain the next position.
In each step, we add a normal distributed random vector to the impulse vector.
Hence we generate series by distorting the first derivative instead of the current position.

\begin{figure}
    \centering
    \includegraphics[width=.49\linewidth]{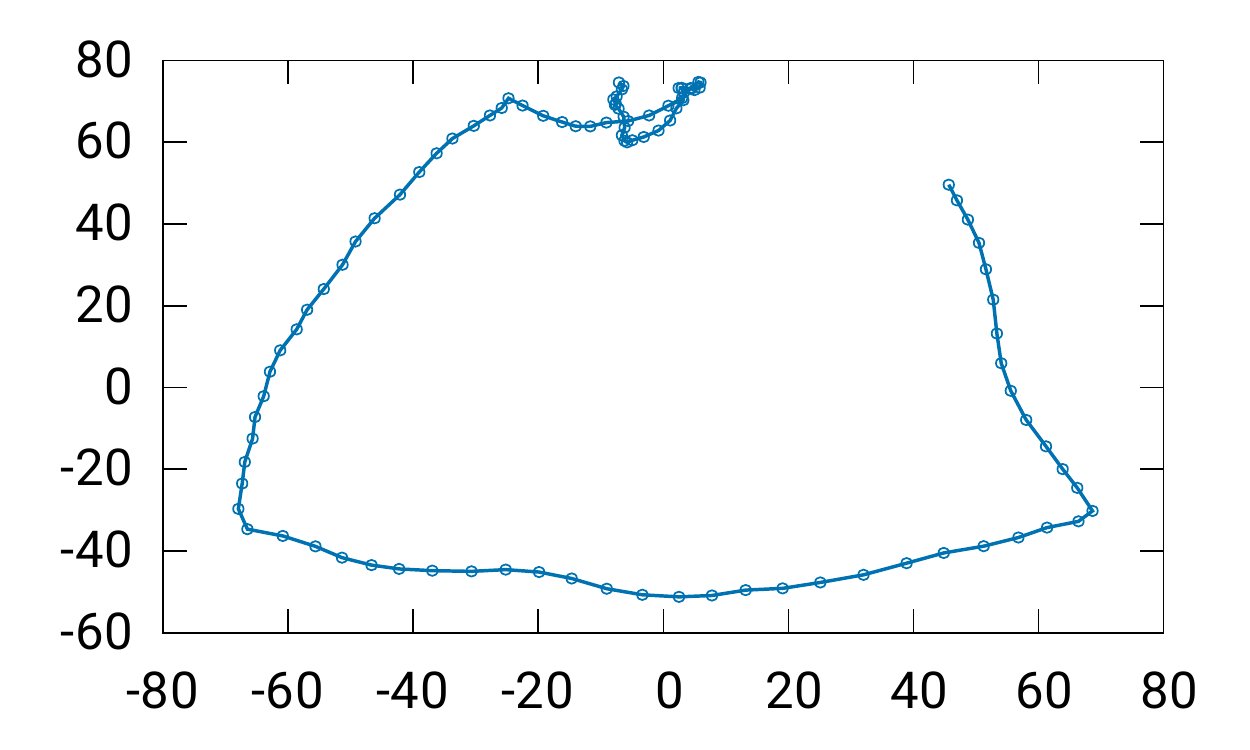}
    \includegraphics[width=.49\linewidth]{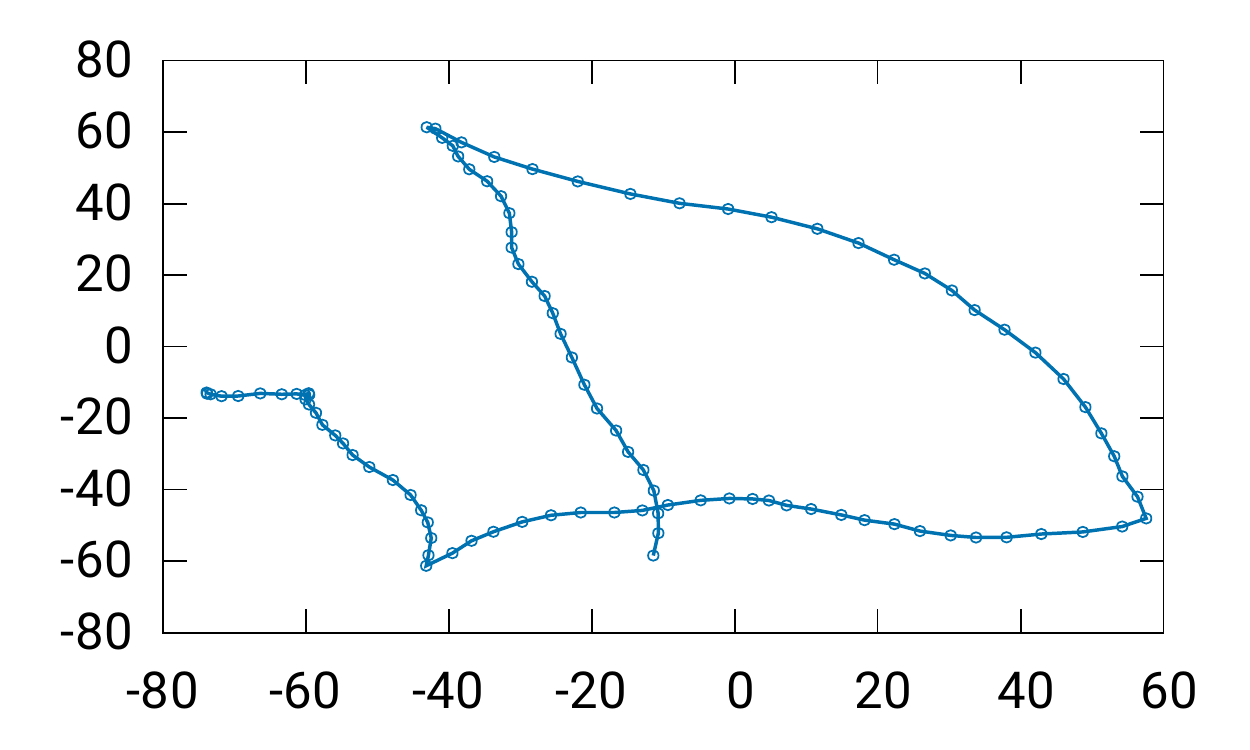}
    \caption{Two examples for a random accelerated motion in 2-dimensional space.}
    \label{fig:velocity}
\end{figure}

To model edges in the generated time series, we restrict the movement to a ball with a constant radius $R$.
When the time series is about to leave the restricted area, we simply let it bounce off the sphere so the time series remains in the interior.
Algorithm~\ref{alg:velocity} provides the pseudocode for data generation and Figure~\ref{fig:velocity} shows an example for a 2-dimensional time series.
In the algorithm, \texttt{uniformBall} returns a uniform distributed vector from the interior of a union spere and \texttt{uniformSphere} returns a uniform distributed vector on a union sphere.

\begin{algorithm}
    \caption{Random Accelerated Motion Generator}
    \begin{lstlisting}
Algorithm: rambase
Input: length $l$, dimensionality $n$, radius $r$
Output: time series $s$

let $s$ be an $n$-dimensional time series of length $l$
$v = (0,\cdots,0)$
$\nu = $ normal$(0, 1)$
$s_0 = $ uniformBall$(r)$
for $i$ from $1$ to $l-1$
    $v = v\,+$ uniformSphere$(r)$
    $s_i = s_{i-1}+v$
    if $\|s_i\|_2 > r$
        // rescale $p$ to stay within the ball with radius $r$
        $s_i = s_i \cdot \frac{r}{\|p\|_2}$
        $v = $ reflect $v$ on sphere at point $p$
return $s$
    \end{lstlisting}
    \label{alg:velocity}
\end{algorithm}

\subsection{Generating Representatives}

We generate representatives of a class corresponding to a base time series by distorting the time series in space and time seperately.

\paragraph{Distortion in Space}

\begin{figure}
    \centering
    \includegraphics[width=.49\linewidth]{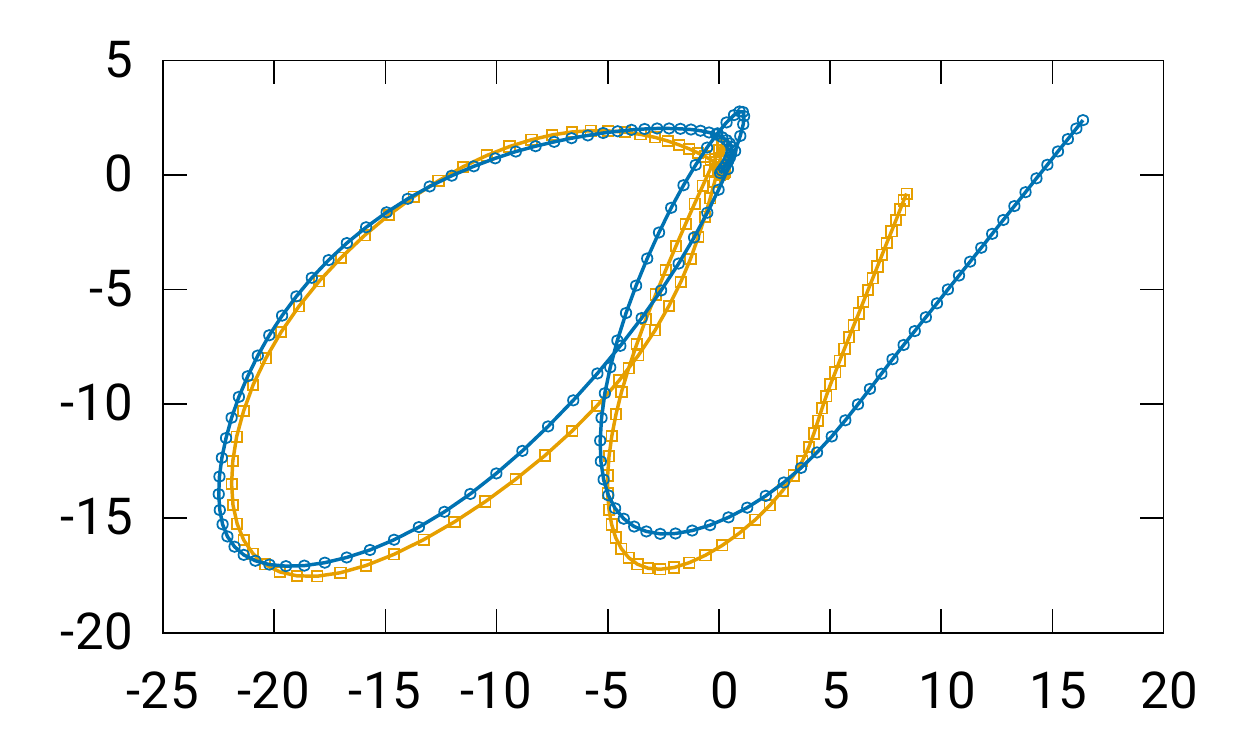}
    \includegraphics[width=.49\linewidth]{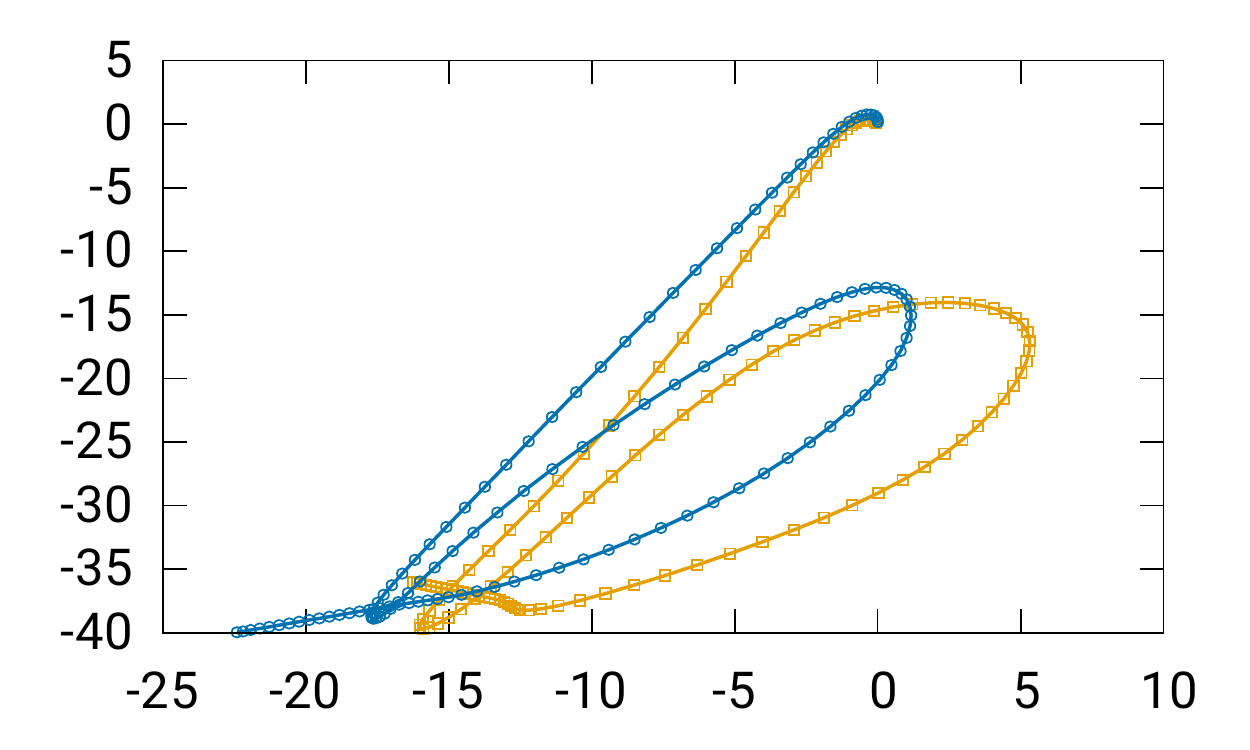}
    \caption{Two time series representatives from two classes of the character trajectories dataset \cite{UCIDatasets}.}
    \label{fig:ct_examples}
\end{figure}

Figure~\ref{fig:ct_examples} shows examples from the character trajectories dataset \cite{UCIDatasets}.
Naturally, the representatives of a class do not match exactly.
We try to imitate this property by adding noise to the first derivative of the time series.
However, to prevent large divergence on long time series, we limit the maximum distance to the base time series by a distortion parameter.

\begin{figure}
    \centering
    \includegraphics[width=.49\linewidth]{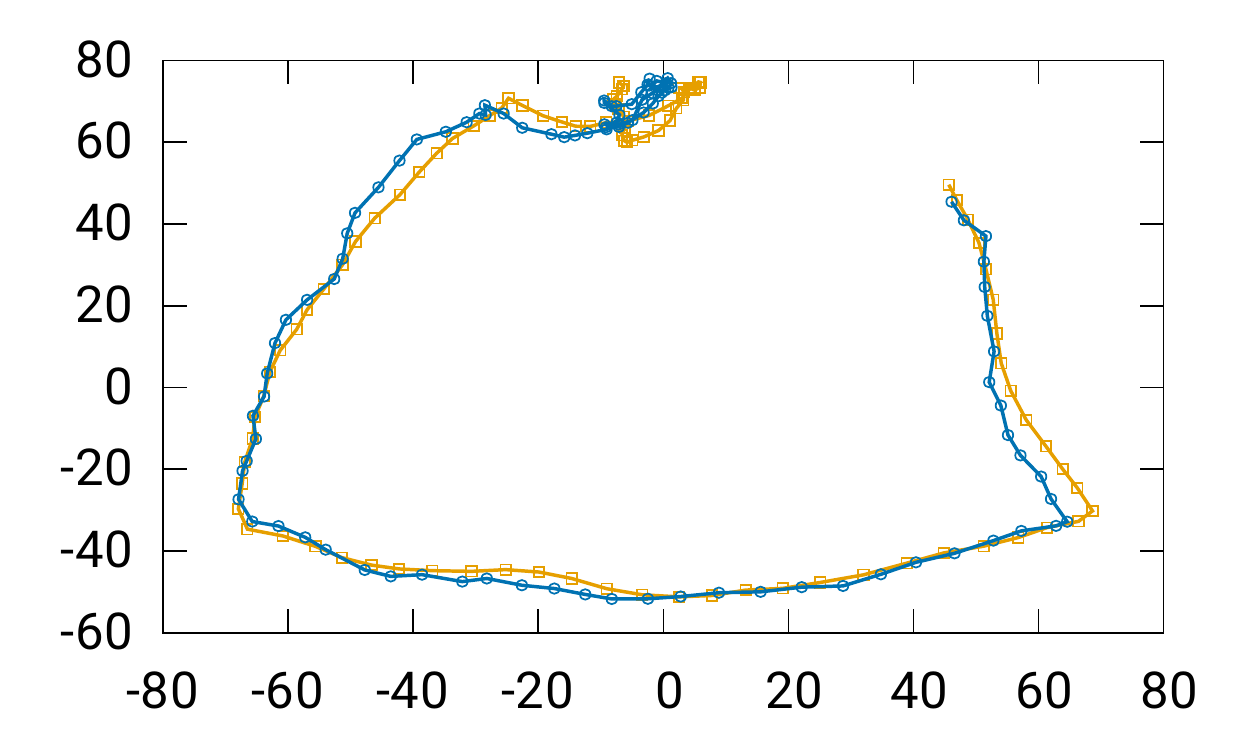}
    \includegraphics[width=.49\linewidth]{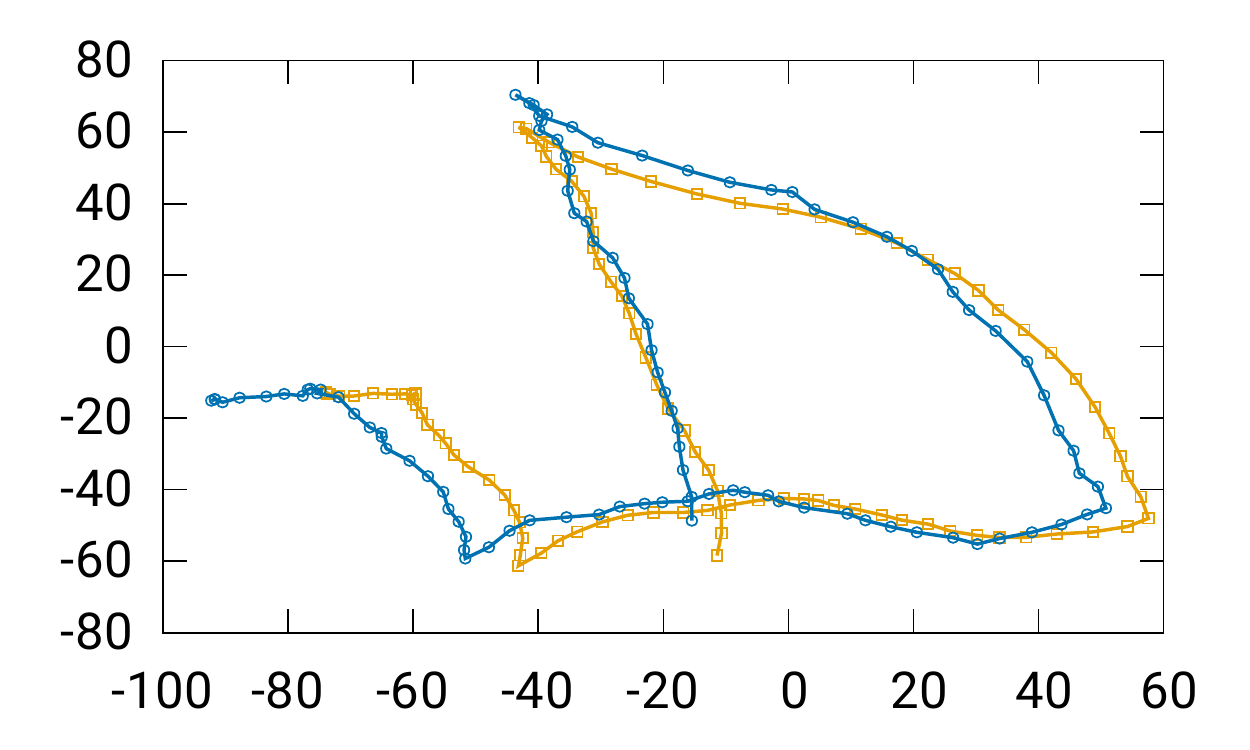}
    \caption{Two time series (yellow) and their distorted derivative (blue); length: $100$; dimensionality: $2$; distortion: $5$ (left) and $25$ (right).}
    \label{fig:ram_spacedistortion}
\end{figure}

Hence, to distort a given time series in space, we add a standard normal distributed random vector to the first derivative of each point analogously to the base time series generator.
This also includes the first point of the time series by assuming that its predecessor is the null point $\textbf{0}=(0,\cdots,0)$.
Figure~\ref{fig:ram_spacedistortion} shows two example time series and a copy for each with distorted derivatives.

Comparing Figure~\ref{fig:ct_examples} and Figure~\ref{fig:ram_spacedistortion} shows indeed that we could imitate the properties from the character trajectories dataset.


\paragraph{Distortion in Time}

\begin{figure}
    \centering
    \includegraphics[width=.49\linewidth]{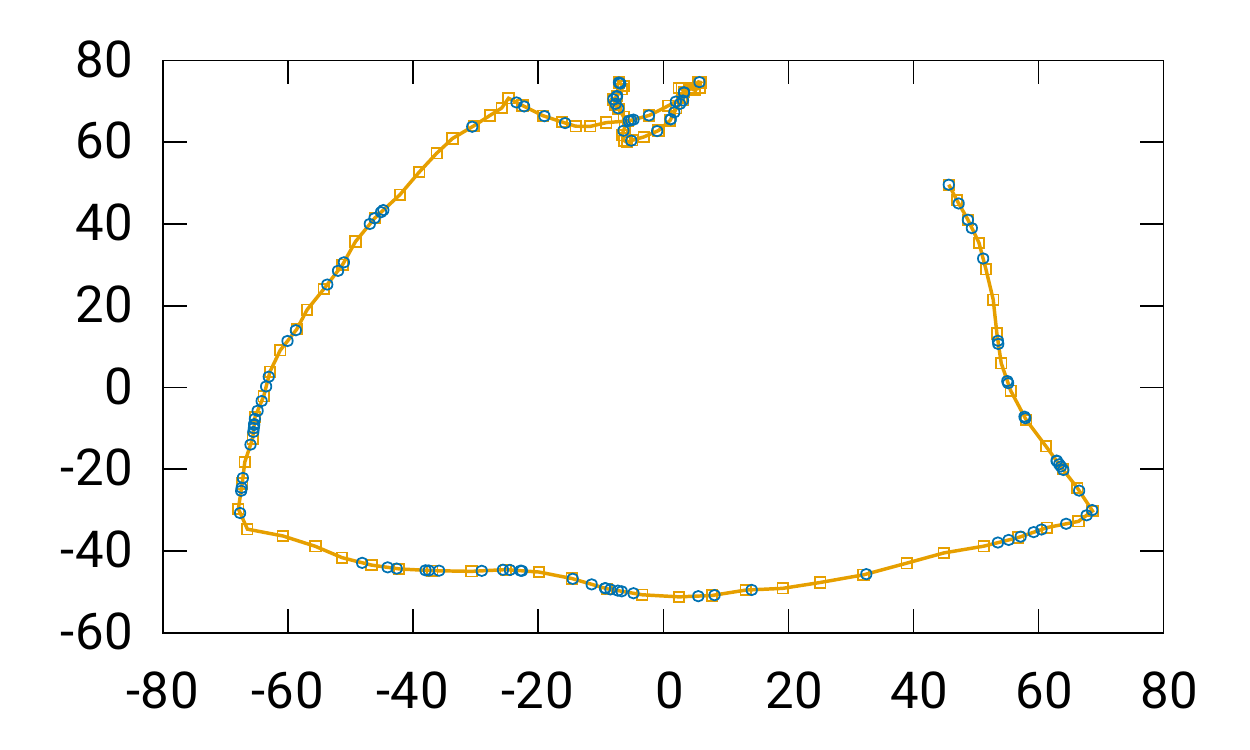}
    \includegraphics[width=.49\linewidth]{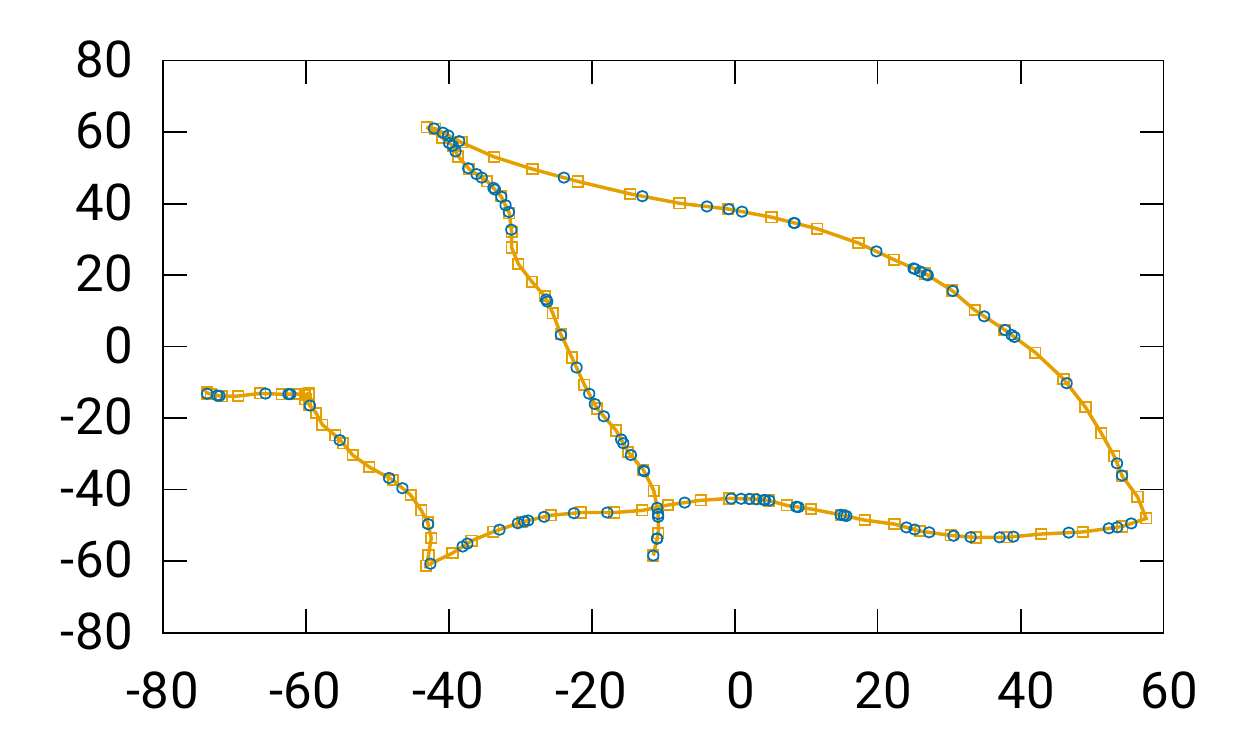}
    \caption{Two time series and their time distorted (blue) versions; length: $100$; dimensionality: $2$.}
    \label{fig:ram_timedistortion}
\end{figure}

In order to apply distortion in time we first interpret the time series as a continuous curve.
Points between two adjacent points of the time series are computed using simple linear interpolation.
Then, we reparameterize the curve in terms of the arc length instead of the time.
Finally, the time distorted time series consists of the first element of the time series, a set of points uniformly distributed on the reparameterized curve, and the last element of the time series.
Algorithm~\ref{alg:timedistortion} provides the pseudocode for the time distortion and Figure~\ref{fig:ram_timedistortion} shows two examples of time distorted time series.

\begin{algorithm}
    \caption{Time Distortion of a Time Series}
    \begin{lstlisting}
Algorithm: timedistortion
Input: time series $s$ of length $L$
Output: time series $\tilde s$

// get arc length up to each point of the time series
$\ell_i=\sum_{j=1}^{i} \|s_j-s_{j-1}\|_2$
// get uniformly distributed values along the complete arc
$t=(0,\ell_{L-1})$
repeat $L-2$ times
    $t=t\,\circ\,$uniform$([0,\ell_{L-1}])$
// interpolate between reparameterized points
$\tilde s = ()$
for $x$ in $sort(t)$
    // find correct index
    $i=\min\left\{ i\mid \ell_i \leq x < \ell_{i+1} \right\}$
    // interpolation parameter
    $u=\frac{x-\ell_i}{\ell_{i+1}-\ell_i}$
    $\tilde s = \tilde s\circ \left( \left( 1-u \right)\cdot s_i + u\cdot s_{i+1} \right)<++>$
return $\tilde s$
    \end{lstlisting}
    \label{alg:timedistortion}
\end{algorithm}

\subsection{Dataset Generator}

As mentioned in the beginning of this section, we use the base time series generator to generate the classes.
The time distortion and space distortion algorithms generate the actual representatives of the classes.

\begin{figure}
    \centering
    \includegraphics[width=.49\linewidth]{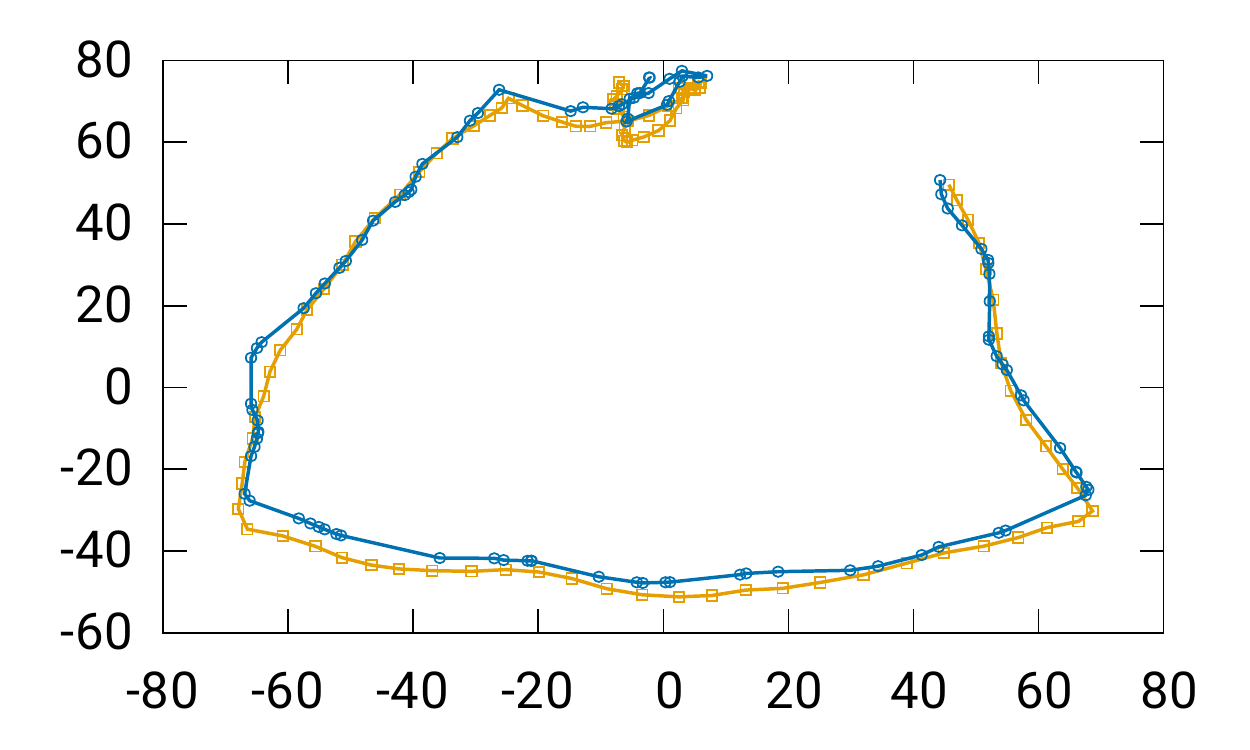}
    \includegraphics[width=.49\linewidth]{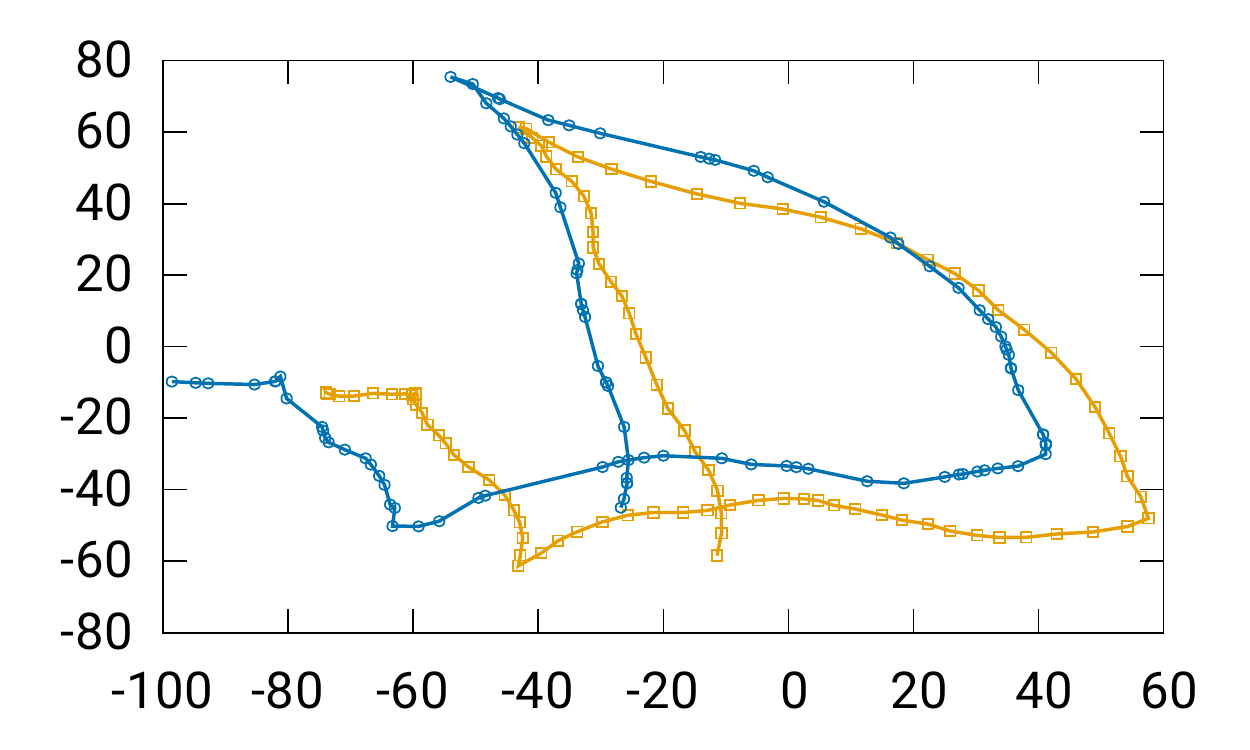}
    \caption{Two base time series and their distorted (blue) versions; length: $100$; dimensionality: $2$; radius: $75$, distortion: $5$ (left) and $25$ (right).}
    \label{fig:ram_full}
\end{figure}

Hence, to generate a dataset $\mathcal D$ with $C$ classes each having $N$ representatives, first call \texttt{rambase} $C$ times to generate $T_{i,0}$ ($1 \leq i \leq C$).
Then, for each $1\leq i\leq C$, call \texttt{timedistortion} and \texttt{spacedistortion} $N$ times to generate $T_{i,j}$ ($1\leq j\leq N$).
The dataset consists of each $T_{i,j}$ for $1\leq i\leq C$ and $1\leq j\leq N$.
\begin{align*}
    \mathcal C &\coloneqq \left\{ T_{i,0} = \texttt{rambase}( L, n, r ) \mid 1\leq i\leq C \right\} \\
    \mathcal R_i &\coloneqq \left\{ T_{i,j} = \texttt{spacedistortion}\big( \right. \\
                 & \quad\quad\quad\quad\quad \left. \texttt{timedistortion}( T_{i,0} ), D\big) \right\} \\
    \mathcal D &\coloneqq \bigcup_{i=1,\ldots,j} \mathcal R_i
\end{align*}
where $n$ is the desired dimensionality, $r$ is the radius of the bounding sphere, $L$ is the desired length of the time series, and $D$ the desired degree of distortion within each class.
Figure~\ref{fig:ram_full} shows two examples of a base time series and their time and space distorted representatives.

\section{Evaluation}
\label{sec:evaluation}

As already emphasized in Section~\ref{sec:introduction}, we propose two dataset generators which produce datasets applicable for classification tasks.
We claimed that they provide a tuning paramater to control the difficulty classification tasks.
In this section, we evaluate, whether our proposed generators (from Section~\ref{sec:cbf} and \ref{sec:ram}) satisfy our claims.


\subsection{Cylinder-Bell-Funnel}

The \cbf{} dataset has no distortion parameter which makes it harder to evaluate the classification strength of time warping distance functions.
However, we can influence the difficulty of the classification task by changing the number of representatives per class.

\begin{figure}[h]
    \centering
    \includegraphics[width=.49\linewidth]{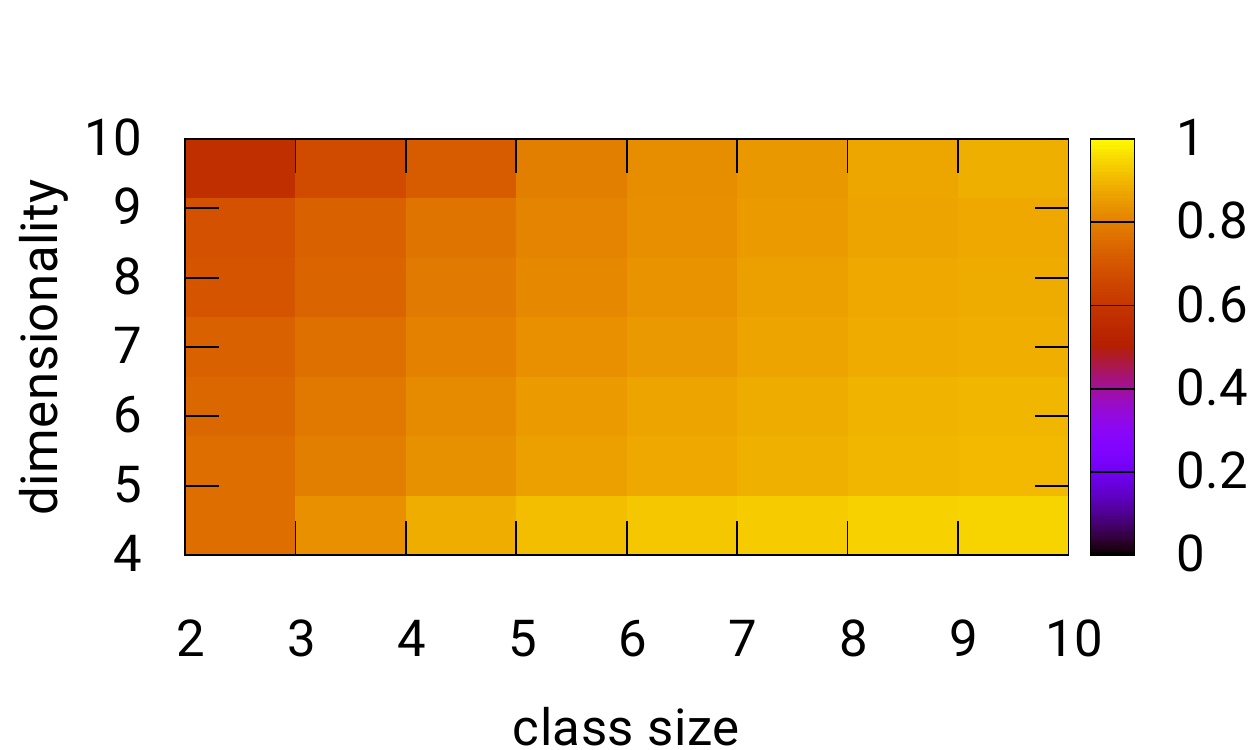}
    \includegraphics[width=.49\linewidth]{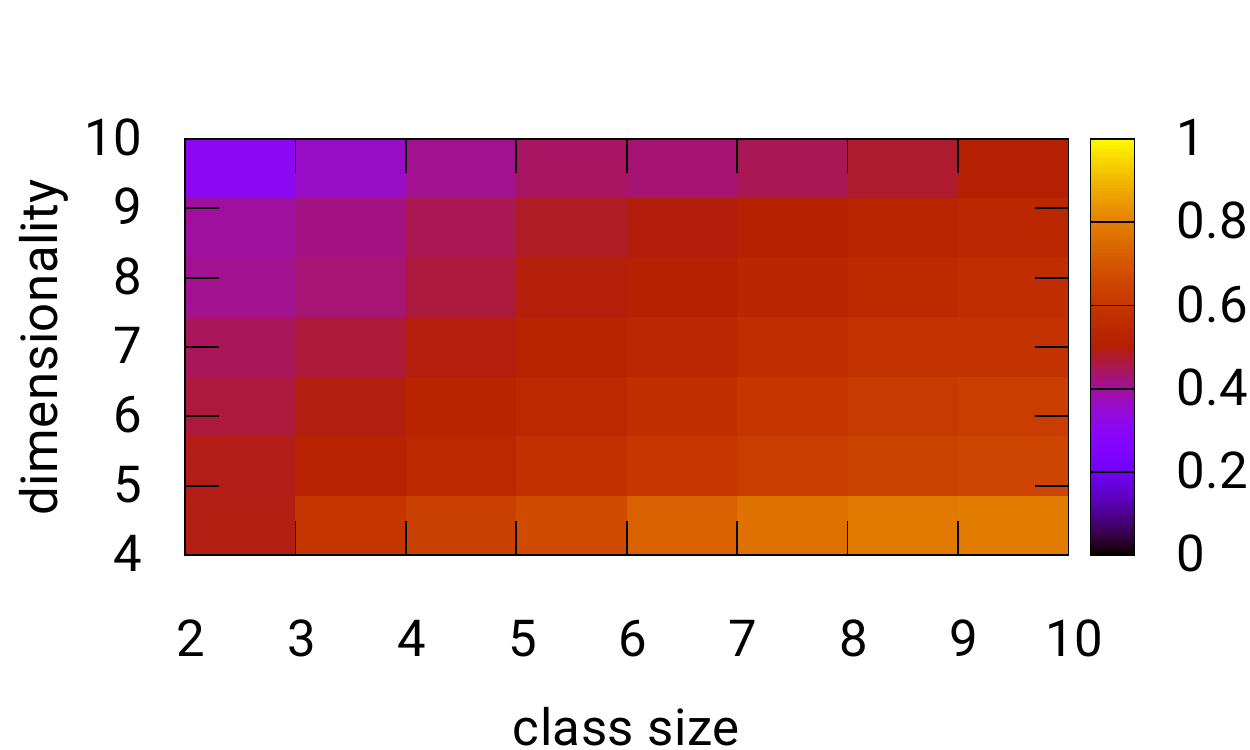}
    \caption{Classification score for \cbf{} dataset using \dtw{} (left) and \ed{} (right); length: 125; number of classes: 27}
    \label{fig:cbf_classsize_dim}
\end{figure}

\begin{figure}[h]
    \centering
    \includegraphics[width=.49\linewidth]{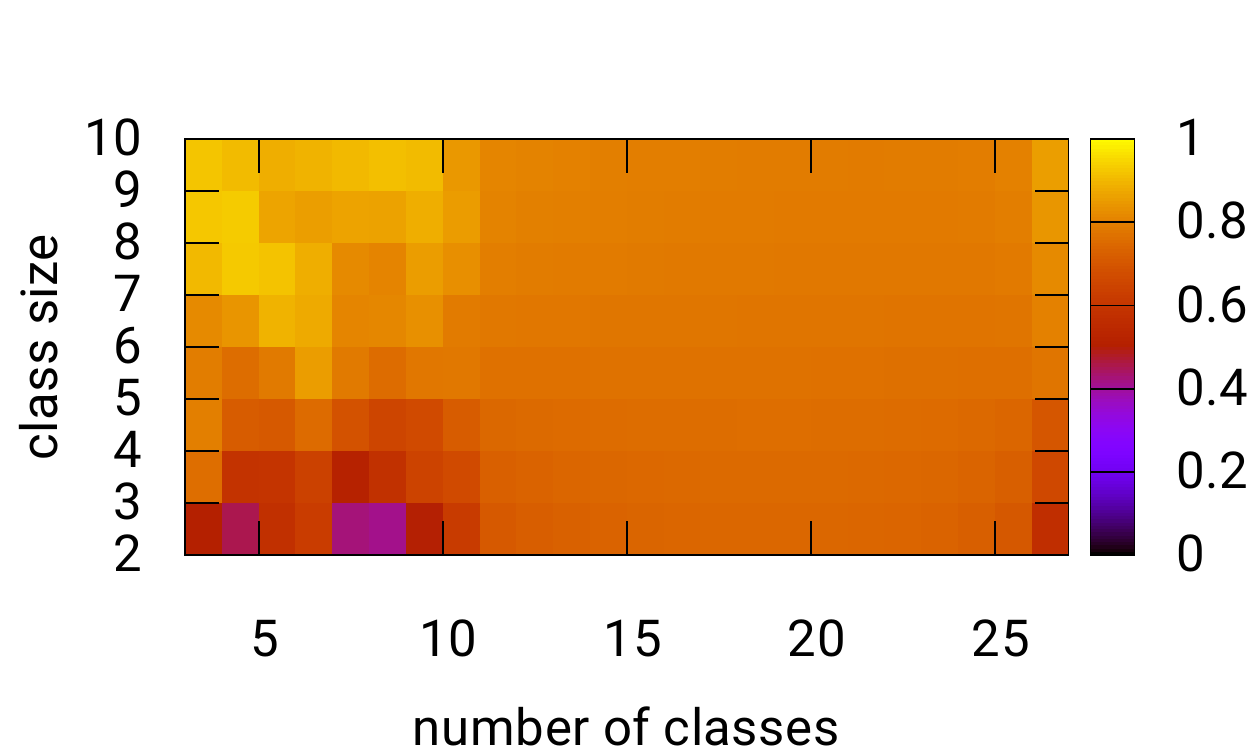}
    \includegraphics[width=.49\linewidth]{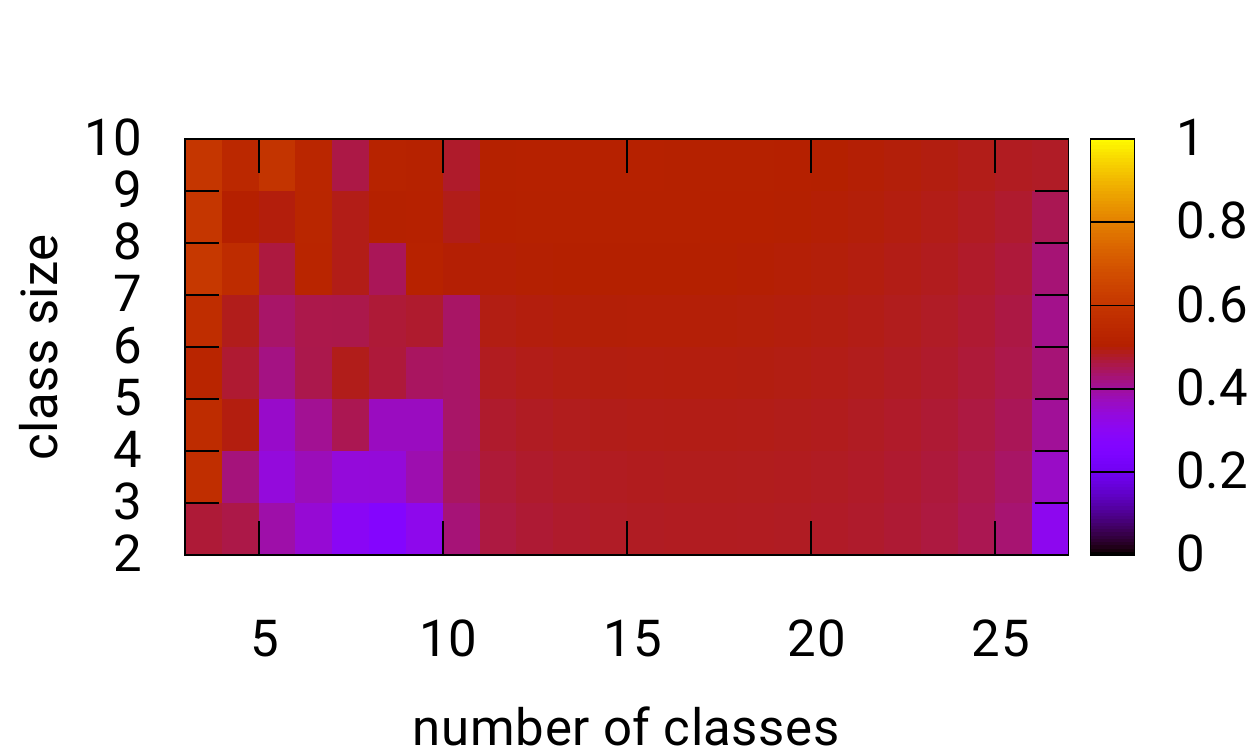}
    \caption{Classification score for \cbf{} dataset using \dtw{} (left) and \ed{} (right); length: 125; dimensionality: 10}
    \label{fig:cbf_numclasses_classsize}
\end{figure}

The heatmap in Figure~\ref{fig:cbf_classsize_dim} shows that for each dimensionality of the \cbf{} dataset the classification score increases with growing class sizes.
However, Figure~\ref{fig:cbf_numclasses_classsize} shows that there are a few cases where the classification score slightly decreases with growing class size.
Hence, we can use the class size to roughly control the difficulty of the classification task with.

Figure~\ref{fig:cbf_classsize_dim} also shows that the classification strength decreases on higher dimensionality.
We could observe that this behaviour does neither depend on the length of the time series nor on the number of classes.

We observe better classification scores using \dtw{} than using the Euclidean distance.
Hence, it appears that time warping distance functions (such as \dtw) are necessary to achieve good classification scores on this dataset.
Whether all time warping distance functions are applicable is out of scope.

\subsection{Random Acceleration Motion}

\begin{figure}[h]
    \centering
    \includegraphics[width=.49\linewidth]{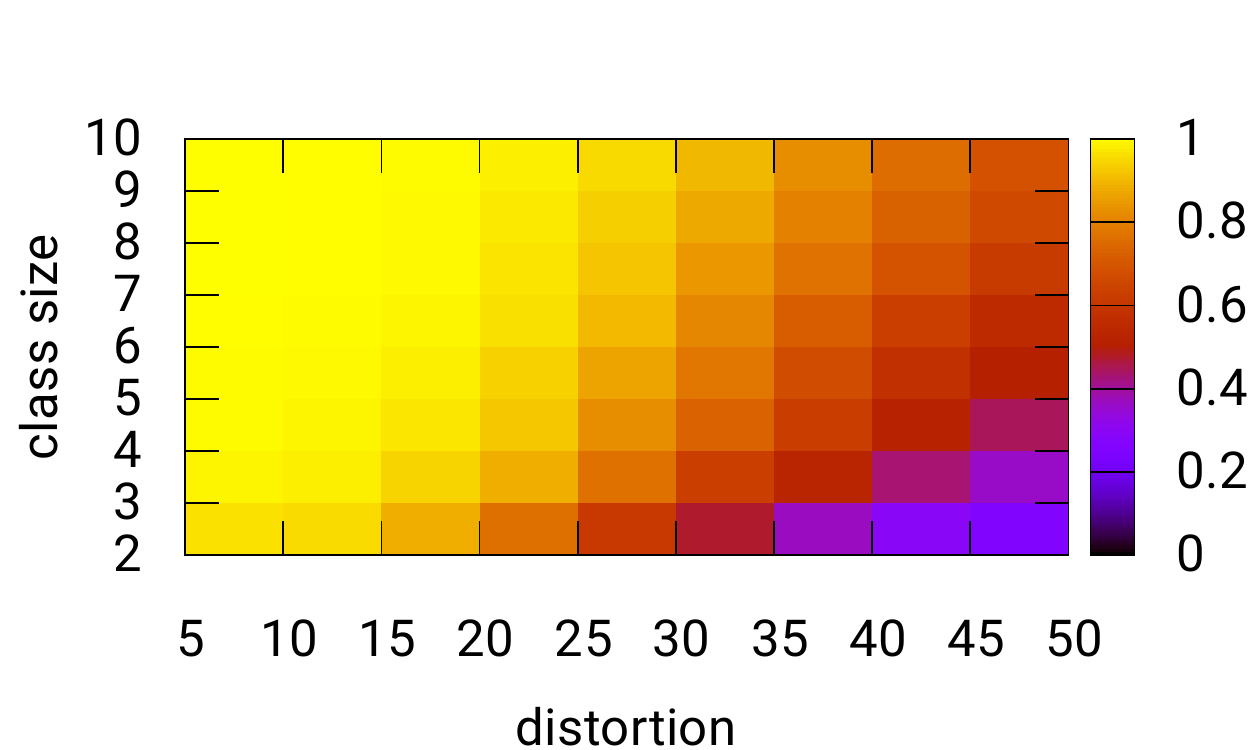}
    \includegraphics[width=.49\linewidth]{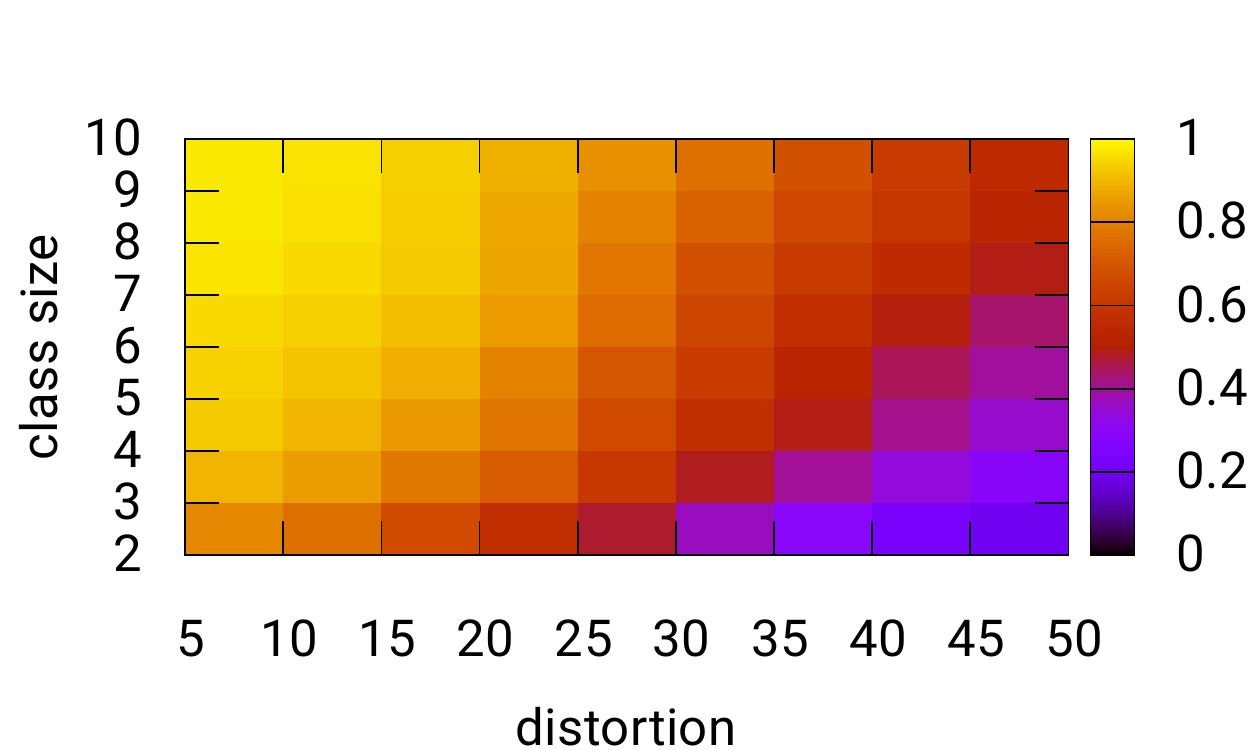}
    \caption{Classification score using \dtw{} (left) and Euclidean distance (right) for an example parameter set: radius $50$, length $100$, dimensionality: $3$, number of classes $200$.}
    \label{fig:ram_dist_classsize_score}
\end{figure}

As mentioned in Section~\ref{sec:ram}, \ram{} generates datasets of multidimensional labeled time series.
Similar to the \cbf{} datasets, Figure~\ref{fig:ram_dist_classsize_score} shows that the classification score increases with growing class size.
Also, this generator has a distortion parameter to control the noisiness of the time series.
This parameter of the \ram{} synthesizer impacts the classification score as expected:
The score decreases with increasing distortion.

\begin{figure}[h]
    \centering
    \includegraphics[width=.49\linewidth]{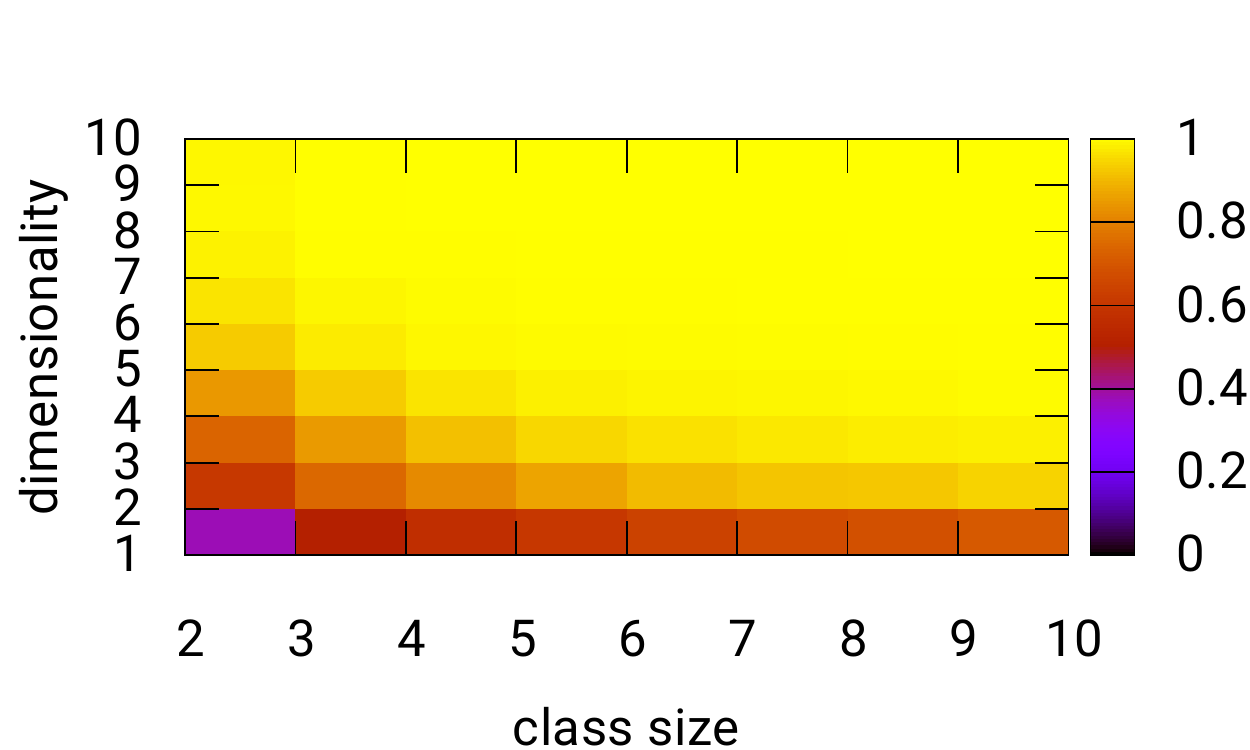}
    \includegraphics[width=.49\linewidth]{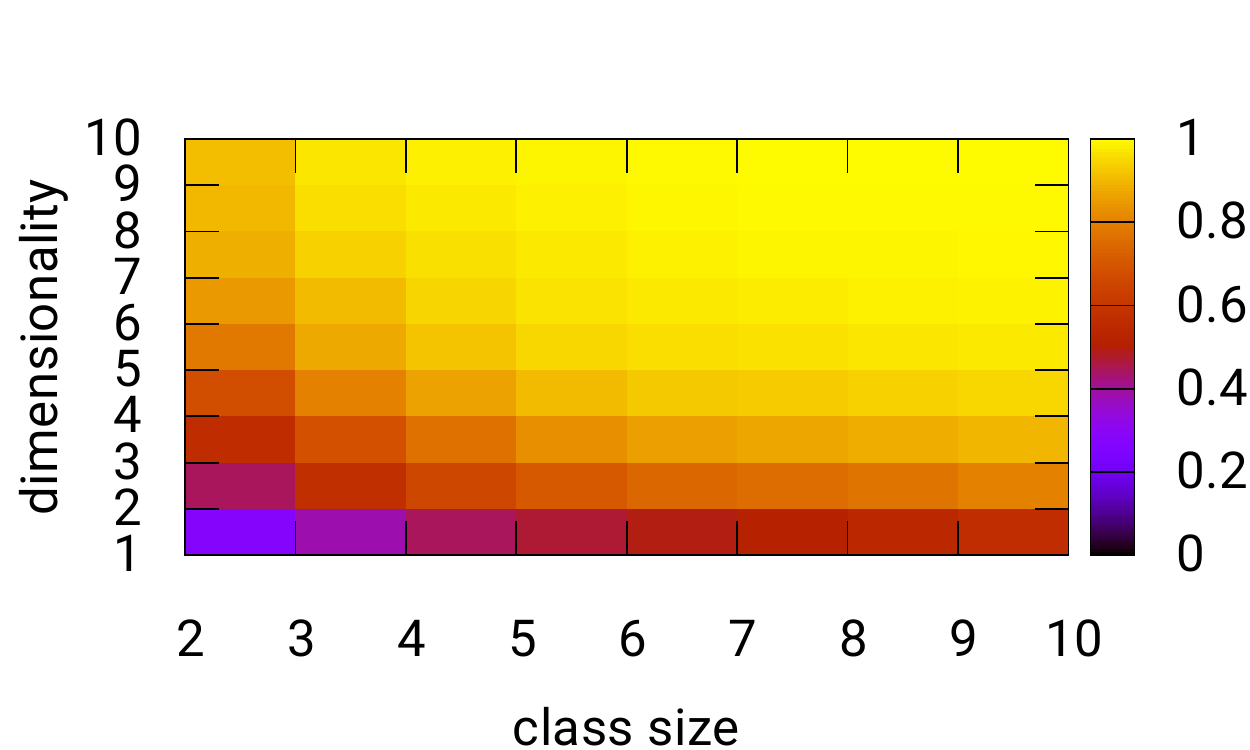}
    \caption{Classification score using \dtw{} (left) and Euclidean distance (right) for an example parameter set: radius $50$, length $100$, number of classes $200$, distortion: $5$.}
    \label{fig:ram_classsize_dim_score}
\end{figure}

Regarding the dimensionality, the \ram{} synthesizer seems to be complementary to the \cbf{} synthesizer, since the classification scores increase with growing dimensionality (c.\,f. Figure~\ref{fig:ram_classsize_dim_score}).

Figure~\ref{fig:ram_dist_classsize_score} furthermore shows that the \dtw{} distance performs better than the Euclidean distance.
Again, it appears time warping distance functions are better suited for solving classification tasks on these datasets.

\section{Conclusion}

We introduced two new dataset generators producing multidimensional labeled time series.
The datasets are applicable for classification tasks using time warping distance functions such as Dynamic Time Warping (\dtw).
Both generators provide parameters adjusting the difficulty of the classification task.
Since the classification scores using \dtw{} increase with growing dimensionality on the \ram{} datasets while decreasing on the \cbf{} dataset, they seem to have some complementary properties.
Thus, both synthesizers seem to be well suited for evaluating classifiers using time warping distance functions in relation to the dimensionality.

\section*{Acknowledgements}
We thank Kevin Trogant for implementing the main parts of the time series synthesizers.
We also thank Jochen Taeschner who we had valuable discussions with and helped us improving the presentation of paper.

\bibliographystyle{plain}
\bibliography{literature}

\end{document}